\documentclass{SCIS2022}
\usepackage{booktabs}       
\usepackage{graphicx}
\usepackage{amsmath,bm}
\usepackage{amsthm}
\newtheorem{prop}{Proposition}
\usepackage{multirow}
\usepackage{makecell}
\usepackage{algorithmic}
\usepackage{setspace}
\usepackage{wrapfig}
\usepackage{xcolor}
\usepackage{stfloats} 
\usepackage{amssymb} 
\DeclareMathOperator*{\argmax}{arg\,max}
\DeclareMathOperator*{\argmin}{arg\,min}
\usepackage{setspace}
\usepackage{wrapfig}
\usepackage{xcolor}

\begin{document}

	\ArticleType{RESEARCH PAPER}
	\Year{2022}
	\Month{}
	\Vol{}
	\No{}
	\DOI{}
	\ArtNo{}
	\ReceiveDate{}
	\ReviseDate{}
	\AcceptDate{}
	\OnlineDate{}
	
	\title{Pareto Adversarial Robustness: Balancing Spatial Robustness and Sensitivity-based Robustness}{Pareto Adversarial Robustness: Balancing Spatial Robustness and Sensitivity-based Robustness}
	

	\author[1]{Ke Sun}{}
	\author[1]{Mingjie Li}{}
	\author[1,2,3]{Zhouchen Lin}{{zlin@pku.edu.cn}}

	\AuthorMark{Author A}
	
	\AuthorCitation{Author A, Author B, Author C, et al}

	
	\address[1]{State Key Lab of General AI, School of Intelligence Science and Technology, Peking University}
	\address[2]{Institute for Artificial Intelligence, Peking University}
	\address[3]{Pazhou Laboratory (Huangpu)}
	
	\abstract{Adversarial robustness, which primarily comprises sensitivity-based robustness and spatial robustness, plays an integral part in achieving robust generalization. In this paper, we endeavor to design strategies to achieve universal adversarial robustness. To achieve this, we first investigate the relatively less-explored realm of spatial robustness. Then, we integrate the existing spatial robustness methods by incorporating both local and global spatial vulnerability into a unified spatial attack and adversarial training approach. Furthermore, we present a comprehensive relationship between natural accuracy, sensitivity-based robustness, and spatial robustness, supported by strong evidence from the perspective of robust representation. Crucially, to reconcile the interplay between the mutual impacts of various robustness components into one unified framework, we incorporate the \textit{Pareto criterion} into the adversarial robustness analysis, yielding a novel strategy called \textit{Pareto Adversarial Training} for achieving universal robustness. The resulting Pareto front, which delineates the set of optimal solutions, provides an optimal balance between natural accuracy and various adversarial robustness. This sheds light on solutions for achieving universal robustness in the future. To the best of our knowledge, we are the first to consider universal adversarial robustness via multi-objective optimization.}
	
	\keywords{Deep Learning, Adversarial Robustness, Reliable Machine Learning, Pareto Optimization, Spatial Robustness}
	
	\maketitle

	
	\section{Introduction}
	Robust generalization serves as an extension of the traditional generalization that is normally achieved via Empirical Risk Minimization for i.i.d. data~\cite{vapnik2015uniform}. However, the test environment could be slightly or dramatically different from the training environment~\cite{krueger2020out} in a robust generalization scenario. Lately, improving the robustness of deep neural networks has been one of the pivotal areas of research, encompassing different threads of research such as adversarial robustness~\cite{goodfellow2014explaining,szegedy2013intriguing}, non-adversarial robustness~\cite{hendrycks2019benchmarking,yin2019fourier}, Bayesian deep learning~\cite{neal2012bayesian,gal2016uncertainty} and causality~\cite{arjovsky2019invariant}. In this paper, we focus on adversarial robustness, where adversarial examples are carefully manipulated by humans to fool machine learning models, e.g., deep neural networks, which could pose serious threats, especially in safety-critical applications. Currently, adversarial training~\cite{goodfellow2014explaining,madry2017towards,ding2018max,ye2021annealing} is regarded as a promising and widely accepted strategy to address this issue.
	
	Like Out-of-Distribution~(OoD) robustness, adversarial robustness also has several aspects~\cite{hendrycks2020many,liu2021model,che2020smgea}, including \textit{sensitivity-based robustness}~\cite{tramer2020fundamental}, i.e., robustness against pixel-wise perturbations (normally within the constraints of an $l_p$ ball), and \textit{spatial robustness}, i.e., robustness against multiple spatial transformations. Computer vision and graphics literature provide a deeper insight into these two aspects, revealing that two main factors determine the appearance of a pictured object~\cite{xiao2018spatially,szeliski2010computer}: (1) lighting and materials, and (2) geometry. Most previous studies on adversarial robustness have focused only on the first factor~\cite{xiao2018spatially} by examining pixel-wise perturbations, e.g., Projected Gradient Descent~(PGD) attacks~\cite{madry2017towards}, assuming that the underlying geometry stays the same after the adversarial perturbation. Only a small proportion of research works have attempted to tackle the less-studied second factor, which includes Flow-based~\cite{xiao2018spatially} and Rotation-Translation~(RT)-based attacks~\cite{engstrom2017rotation,engstrom2019exploring}. 
	
	However, it is crucial to consider spatial robustness for achieving universal robustness, the ultimate objective of robust generalization. One of the most important reasons is that sensitivity-based robustness, which is generally based on the $l_p$-distance, is not sufficient to maintain perceptual similarity~\cite{sharif2018suitability,engstrom2017rotation,engstrom2019exploring,xiao2018spatially}. Specifically, although \textit{spatial attacks or geometric transformations} result in small perceptual differences, they yield large $l_p$ distances.
	
	A clear relationship between accuracy, sensitivity-based and spatial robustness is the key to achieving universal adversarial robustness. While the trade-off between \textit{sensitivity-based robustness} and accuracy has been revealed by several studies~\cite{zhang2019theoretically,tsipras2018robustness,raghunathan2020understanding}, the comprehensive relationships among \textit{spatial robustness} and them are still unclear. Although previous studies~\cite{tramer2019adversarial,kamath2020invariance} have explored this issue, they only focused on Rotation-Translation spatial robustness and did not consider Flow-based spatial robustness~\cite{xiao2018spatially,zhang2019joint}. Surprisingly, we find that Flow-based spatial robustness presents a relationship contrary to the one revealed previously, making the previous conclusion less reliable.
	
	Based on this important finding, we start our exploration of clearer relationships between different robustnesses, and we eventually harmonize the conflicting relationships within them by leveraging the Pareto criterion~\cite{kim2005adaptive,kim2006adaptive,zeleny2012multiple}, thus achieving an optimal balanced universal robustness. A recent study~\cite{raghunathan2020understanding} attributes the conflicting relationships among the various robustnesses to overparametrization, while we uncovered it from the perspective of different shape-biased representations. Another report~\cite{wang2020once} examined the trade-off in the inference time, while we target more comprehensive relationships between different robustnesses with a different methodology.
	
	In this paper, we first try to gain deeper insights into the robustness relationships by investigating the two main spatial robustness branches, i.e., Flow-based spatial attack~\cite{xiao2018spatially} and Rotation-Translation~(RT) attack~\cite{engstrom2019exploring}. After revealing their impact
	on local and global spatial sensitivity, we propose integrated spatial attack and spatial adversarial training, which can incorporate comprehensive spatial vulnerabilities or robustness. Based on this understanding, we present a comprehensive relationship among the accuracy, sensitivity-based robustness, and the two branches of spatial robustness by investigating their different saliency maps from the perspectives of shape-bias, sparse or dense representation. It turns out that while the relationship between sensitivity-based and RT robustness is a fundamental trade-off, sensitivity-based and Flow-based spatial robustness are highly correlated, providing a vital supplementary for previous conclusions. Thus, comprehensive relationships between accuracy and the various robustnesses are not pure trade-offs, motivating us to introduce the Pareto criterion~\cite{kim2005adaptive,kim2006adaptive,zeleny2012multiple}, the general multi-objective optimization principle, into the universal adversarial robustness analysis. The Pareto criterion enables an optimal balance between the interplay of natural accuracy and the different adversarial robustnesses, leading to universal adversarial robustness in a Pareto manner. By incorporating a two-moment term that can capture the interaction between loss of accuracy and different robustnesses, we propose a bi-level optimization framework called \textit{Pareto Adversarial Training}. The resulting Pareto front provides a set of optimal solutions that can balance perfectly all the relationships under consideration, outperforming other existing strategies.

	Our contributions can be summarized as follows:
	
	\begin{itemize}
		\item We reveal the existence of both local and global spatial robustness and propose integrated spatial attack and spatial adversarial training, incorporating comprehensive spatial vulnerabilities.
		
		\item We present comprehensive relationships among accuracy, sensitivity-based, and different spatial robustnesses, supported by strong and intuitive evidence from the perspective of robust representation.
		
		\item We incorporate the Pareto criterion into adversarial robustness analysis, and the resulting Pareto Adversarial Training can optimally balance multiple adversarial robustness, yielding universal adversarial robustness.
	\end{itemize}

	\section{Local and Global Spatial Robustness}\label{sec:spatial}
	
	To present the comprehensive relationships between accuracy and different adversarial robustnesses, we first provide a fine-grained understanding of spatial robustness. We summarize several studies about spatial robustness~\cite{engstrom2017rotation,engstrom2019exploring,xiao2018spatially,zhang2019joint,tramer2019adversarial,kamath2020invariance} into two major branches: (1) Flow-based Attacks, and (2) Rotation-Translation~(RT) Attacks. In particular, we find that the former mainly focuses on the local spatial vulnerability while the latter tends to capture the global spatial sensitivity. Based on this finding, integrated spatial attack and spatial adversarial training are proposed.

	\begin{figure*}[t!]
		\centering
		\includegraphics[width=1.0\textwidth,trim=0 270 20 105,clip]{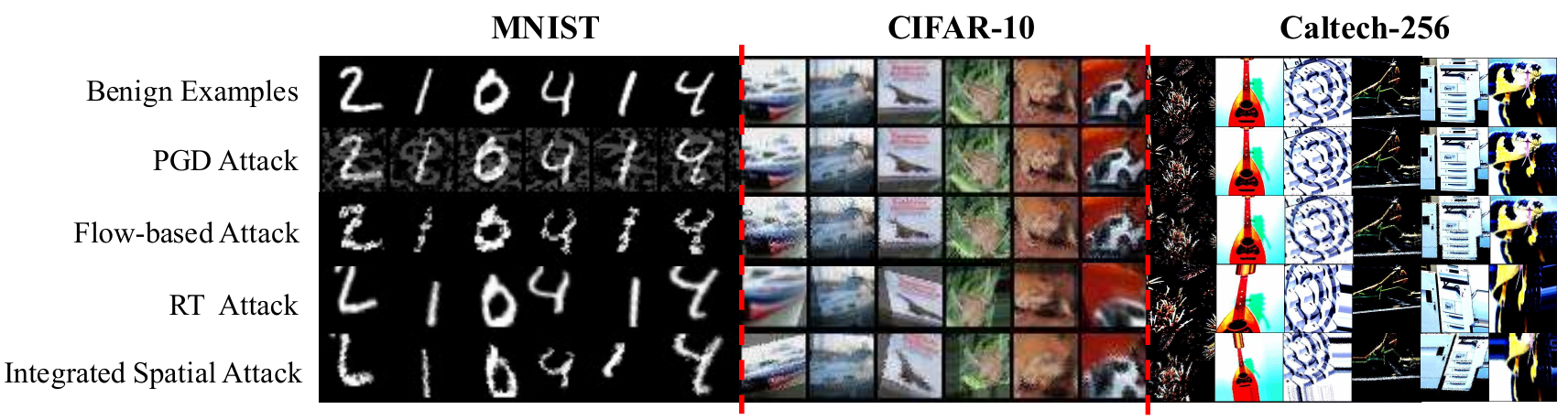}
		\caption{Visualization of Flow-based, RT and Our Integrated Spatial adversarial examples on MNIST, CIFAR-$10$ and Caltech-$256$. More images and detailed discussions are provided in \ref{appendix:moreimages}.}
		\label{figure_1examples}
	\end{figure*}
	
	\subsection{Local Spatial Robustness: Flow-based Attacks}
	The most representative Flow-based Attack is the Spatial Transformed Attack~\cite{xiao2018spatially}, wherein a differentiable flow vector $w_F=(\Delta \mu, \Delta v)$ is introduced in the 2D coordinates $(\mu, v)$ to craft adversarial spatial transformation. The vanilla-targeted Flow-based attack~\cite{xiao2018spatially} follows the optimization manner ($\kappa=0$):
	\begin{equation}\label{eq_flow}
		\begin{aligned} 
			w_F^{*} = \argmin_{w_F} \max_{i \neq t} f_{\theta}^{i}(x_{w_F}) - f_{\theta}^{t}(x_{w_F}) + \tau \mathcal{L}_{f l o w}(w_F),
		\end{aligned} 
	\end{equation}
	where $f_{\theta}(x)=\left(f_{\theta}^{1}(x), \ldots, f_{\theta}^{K}(x)\right)$ is the classifier in the $K$-classification task. $x_{w_F}$ is a Flow-based adversarial example parameterized by the flow vector $w_F$. $\mathcal{L}_{f l o w}$, which measures the local smoothness of the spatial transformation balanced by $\tau$.
	
	Interestingly, our empirical study shown in the left part of Figure~\ref{figure_1examples} suggests that the Flow-based attack tends to yield local permutations among pixels in some specific regions, irrespective of the option of $\tau$, rather than a  global spatial transformation based on their \textit{shapes}. Our analysis indicates that this phenomenon is due to two factors: 1) Local permutations, especially in regions where colors of pixels change dramatically, are already sufficiently sensitive to manipulations, as demonstrated by our empirical results shown in Figure~\ref{figure_1examples}. 2) The manner of optimization does not incorporate any sort of \textit{shape transformation} information, e.g., a parametric equation of rotation, as opposed to the vanilla Rotation-Translation attack, which we present in the following. Therefore, we conclude that Flow-based attacks tend to capture the local spatial vulnerability. Further, to design the integrated spatial attack, we transform Eq~\ref{eq_flow} into its untargeted version under cross-entropy loss with flow vector bounded by an $\epsilon_{F}$-ball:
	\begin{equation}\label{eq_flow_ce}
		\begin{aligned} 
			w^{*}_F = \argmax_{w_F} \mathcal{L}^{\text{CE}}_{\theta} (x_{w_F},y)\ \ s.t. \ \Vert w_F \Vert \le \epsilon_{F},
		\end{aligned} 
	\end{equation}
	where $\mathcal{L}_{\theta}^{\mathrm{CE}}(x, y)=\log \sum_{j} \exp \left(f_{\theta}^{j}(x)\right)-f_{\theta}^{y}(x)$. To maintain a uniform optimization form in our integrated spatial attack, we replace local smoothness term  $\mathcal{L}_{f l o w}$ in Eq.~\ref{eq_flow} with our familiar $l_p$ constraint and leverage the cross-entropy loss instead of the $max$ operation as suggested in~\cite{carlini2017towards}. Proposition~\ref{prop:flow} reveals the correlation between the two losses, indicating that the smooth approximation version of $max$ operation in Eq.~\ref{eq_flow}, denoted as $\mathcal{L}^{S}_{\theta}$, has a parallel updating direction with cross-entropy loss related to $w_F$. Proof can be found in \ref{appendix:proof_flow}.
	
	\begin{prop}\label{prop:flow}
		Consider $\mathcal{L}^{S}_{\theta}(x, y)=\log \sum_{i \neq y} \exp \left(f_{\theta}^{i}(x)\right)-f_{\theta}^{y}(x)$ as the smooth version loss of Eq.~\ref{eq_flow} without a local smoothness term. For a fixed $(x_{w_F},y)$ and $\theta$, we have 	
		\begin{equation}\begin{aligned}
				\nabla_{w_F} \mathcal{L}_{\theta}^{\text{CE}}(x_{w_F}, y)=r(x_{w_F}, y) \nabla_{w_F} \mathcal{L}_{\theta}^{S}(x_{w_F}, y),
		\end{aligned}\end{equation}
		where $r(x_{w_F}, y)= \sum_{i \neq y} \exp \left(f_{\theta}^{i}(x_{w_F})\right) / \sum_{i} \exp \left(f_{\theta}^{i}(x_{w_F})\right)$.
	\end{prop}

	\subsection{Global Spatial Robustness: Rotation-Translation Attacks} 
	The original Rotation-Translation attack~\cite{engstrom2017rotation,engstrom2019exploring} applies parametric equation constraints on 2D coordinates, thus capturing the global spatial information:	
	\begin{equation}\label{eq_RT}
		\begin{aligned} 
			\left[\begin{array}{l}
				u^{\prime} \\
				v^{\prime}
			\end{array}\right]=\left[\begin{array}{cc}
				\cos \theta & -\sin \theta \\
				\sin \theta & \cos \theta
			\end{array}\right] \left[\begin{array}{l}
				u \\
				v
			\end{array}\right]+\left[\begin{array}{l}
				\delta u \\
				\delta v
			\end{array}\right].
		\end{aligned} 
	\end{equation}
	To design a generic spatial transformation matrix that can simultaneously consider rotation, translation, cropping, and scaling, we re-parameterize the transform matrix as a generic 6-dimensional affine transformation one, inspired by Spatial Transformer Networks~\cite{jaderberg2015spatial}:	
	\begin{equation}\label{eq_RT_new}
		\begin{aligned} 
			\left[\begin{array}{l}
				u^{\prime} \\
				v^{\prime}
			\end{array}\right]= (\left[\begin{array}{ccc}
				1 & 0 & 0 \\
				0 & 1 & 0
			\end{array}\right] +	
			\left[\begin{array}{ccc}
				w_{RT}^{11} & w_{RT}^{12} & w_{RT}^{13} \\
				w_{RT}^{21} & w_{RT}^{22} & w_{RT}^{23}
			\end{array}\right]) \left[\begin{array}{l}
				u \\
				v \\
				1
			\end{array}\right],
		\end{aligned} 
	\end{equation}
	where we denote $A_{w_{RT}}$ as the generic 6-dimensional affine transformation matrix, in which each entry of $w_{RT}$ indicates the increment in different spatial aspects. For example, $(w_{RT}^{13},w_{RT}^{23})$ determines translation. Finally, the optimization form of the resulting generic and differentiable RT-based attack bounded by $\epsilon_{RT}$-ball is expressed as:	
	\begin{equation}\label{eq_RT_optimization}
		\begin{aligned} 
			w^{*}_{RT} = \argmax_{w_{RT}} \mathcal{L}^{\text{CE}}_{\theta} (x_{w_{RT}},y) \ \ s.t. \ \Vert w_{RT} \Vert \le \epsilon_{RT}.
		\end{aligned} 
	\end{equation}

	\subsection{Integrated Spatial Attack} The key to achieving integrated spatial robustness is to design an integrated parameterized sampling grid $\mathcal{T}_{w_{RT},w_F}\left(G\right)$ that can wrap the regular grid with both flow and affine transformation, where $G$ is the generated grid. We show our integrated approach as shown below:
	\begin{equation}\label{eq_integrated}
		\begin{aligned} 
			\mathcal{T}_{w_{RT},w_F}\left(G\right) & = A_{w_{RT}}\left[\begin{array}{l}
				u \\
				v \\
				1
			\end{array}\right] + 
			w_F, \\
			x^{adv} &= \mathcal{T}_{w_{RT},w_F}\left(G\right) \circ x.
		\end{aligned} 
	\end{equation}
	Then we sample new $x^{adv}$ by $\mathcal{T}_{w_{RT},w_F}\left(G\right)$ via the differentiable bilinear interpolation~\cite{jaderberg2015spatial}. Note that $w_F$ has the same dimensions as the grid $G$, which are different from the impact of two-dimensional translation parameters in $w_{RT}$. Then the final loss function of the integrated spatial attack can be presented as:
	\begin{equation}\label{eq_integrated_optimization}
		\begin{aligned} 
			w^{*} = \argmax_{w} \mathcal{L}^{\text{CE}}_{\theta} (x + \eta_{w},y), \ \ s.t. \ \Vert w \Vert \le \epsilon,
		\end{aligned} 
	\end{equation}
	where $\eta_{w}$ is the crafted integrated spatial perturbation parameterized by $w=[w_{F},w_{RT}]^T$, simultaneously considering both Flow-based and RT spatial sensitivity. Note that $\eta_w$ itself does not necessarily satisfy the $l_p$ constraint directly. For the implementation, we follow the PGD procedure~\cite{madry2017towards}, a common practice in sensitivity-based attacks. We consider the infinity norm of $w$ and different learning rates for the two types of spatial robustness. Therefore, the updating rule of $w$ in each iteration is:
	\begin{equation}\label{eq_integrated_PGD}
		\begin{aligned} 
			\left[\begin{array}{l}
				\bar{w}^{t+1}_F \\
				\bar{w}^{t+1}_{RT}
			\end{array}\right]  & = \left[\begin{array}{l}
				w^{t}_F \\
				w^{t}_{RT}
			\end{array}\right] + \left[\begin{array}{l}
				\alpha_F \\
				\alpha_{RT}
			\end{array}\right] \text{sign}(\nabla_{w} \mathcal{L}^{\text{CE}}_{\theta}(x^t_{w^t},y)), \\
			\left[\begin{array}{l}
				w^{t+1}_F \\
				w^{t+1}_{RT}
			\end{array}\right]&=\text{clip}_{\epsilon}(\left[\begin{array}{l}
				\bar{w}^{t+1}_F \\
				\bar{w}^{t+1}_{RT}
			\end{array}\right]),\\
			x^{t+1}_{w^{t+1}} &= \mathcal{T}_{w^{t+1}}\left(G\right) \circ x,
		\end{aligned} 
	\end{equation}
	where $w^{t+1}=[w^{t+1}_F, w^{t+1}_{RT}]^T$ is element-wisely clipped from $\bar{w}^{t+1}$  by $\epsilon=[\boldsymbol{\epsilon}_{F},\boldsymbol{\epsilon}_{RT}]^T$. From Figure~\ref{figure_1examples}, we can observe that our Integrated Spatial Attack can construct both local and global spatial transformations on images. Thus, it can simultaneously yield local pixel-wise permutations and global shape transformations.

	\begin{figure*}[t!]
		\centering
		\includegraphics[width=0.95\textwidth,trim=10 235 10 150,clip]{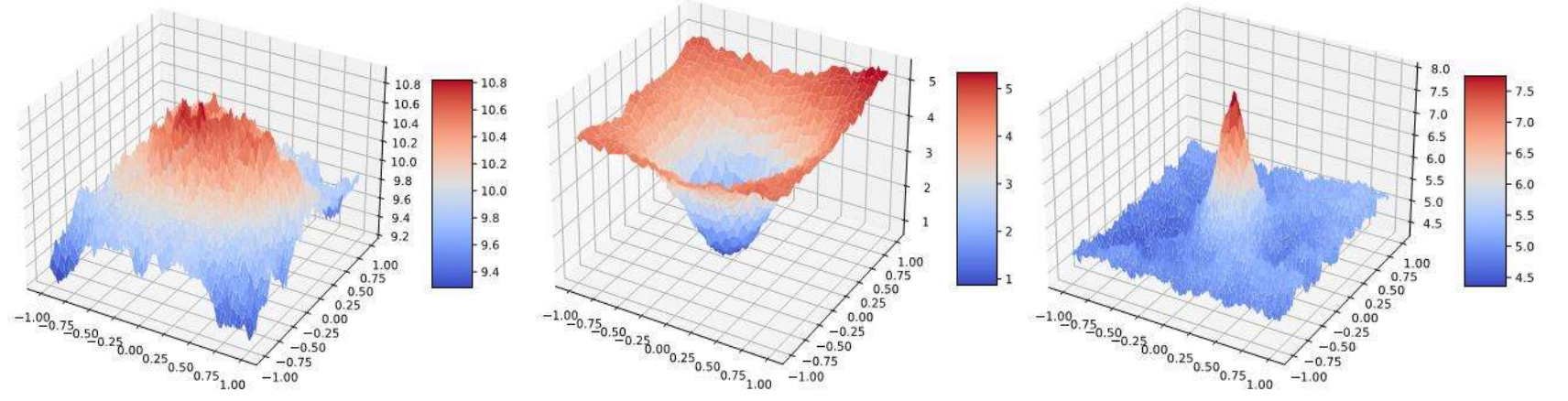}
		\caption{Loss landscape of Integrated Spatial Attack on CIFAR-$10$. (Left) A distant view of loss landscape w.r.t $w$ before the optimization in Eq.~\ref{eq_integrated_optimization}. (Middle) A close view before the optimization shows a highly convex surface near the initialization point. (Right) The loss landscape around the maxima $w^*$ after the optimization in Eq.~\ref{eq_integrated_optimization}.}
		\label{figure_2landscape}
	\end{figure*}

	Then, we visualize the loss surface under this Integrated Spatial Attack leveraging ``filter normalization''~\cite{li2018visualizing} as illustrated in Figure~\ref{figure_2landscape}. We strictly follow the implementation from \cite{li2018visualizing} to achieve the desired visualization of the loss landscape of our integrated adversarial attacks for all the differentiable parameters $w$. Specifically, we view $w_{F}$ and $w_{RT}$ as two parameterized filters, which is analogous to the ``filter normalization?? technique proposed by \cite{li2018visualizing}. In the left part of Figure~\ref{figure_2landscape}, we adjust the initialization of the variance of $w$, which then can provide a distant view of loss landscape before the optimization in Eq.~\ref{eq_integrated_optimization}. It exhibits a highly regular loss landscape, and its non-concavity w.r.t. only rotation and translation ~\cite{engstrom2019exploring} has been tremendously improved. In the middle of Figure~\ref{figure_2landscape}, we then provide a closer view of the loss landscape before the optimization. It shows a highly convex surface around the $w$ to be optimized, facilitating the following optimization. In the right part of Figure~\ref{figure_2landscape}, we also present the loss landscape around the maxima $w^*$ after the optimization in Eq.~\ref{eq_integrated_optimization} of our integrated spatial attack, exhibiting a highly concave surface as well. In summary, the highly non-concave loss landscape concerning only rotation and translation raised by \cite{engstrom2019exploring} has been largely alleviated by considering both local and global spatial vulnerabilities. This integrated form smooths the optimization process, which guarantees the efficacy of our Integrated Spatial Attack.
	
	\subsection{Spatial Adversarial Training}\label{spatialAT}
	
	As Eq.~\ref{eq_integrated_PGD} incorporates local and global spatial robustness simultaneously, it is natural to leverage it to construct Spatial Adversarial Training, which we deploy in Experiment~\ref{experiment:paretofront}.
	
	\begin{figure*}[t!]
		\centering
		\includegraphics[width=1.0\textwidth,trim=0 0 0 10,clip]{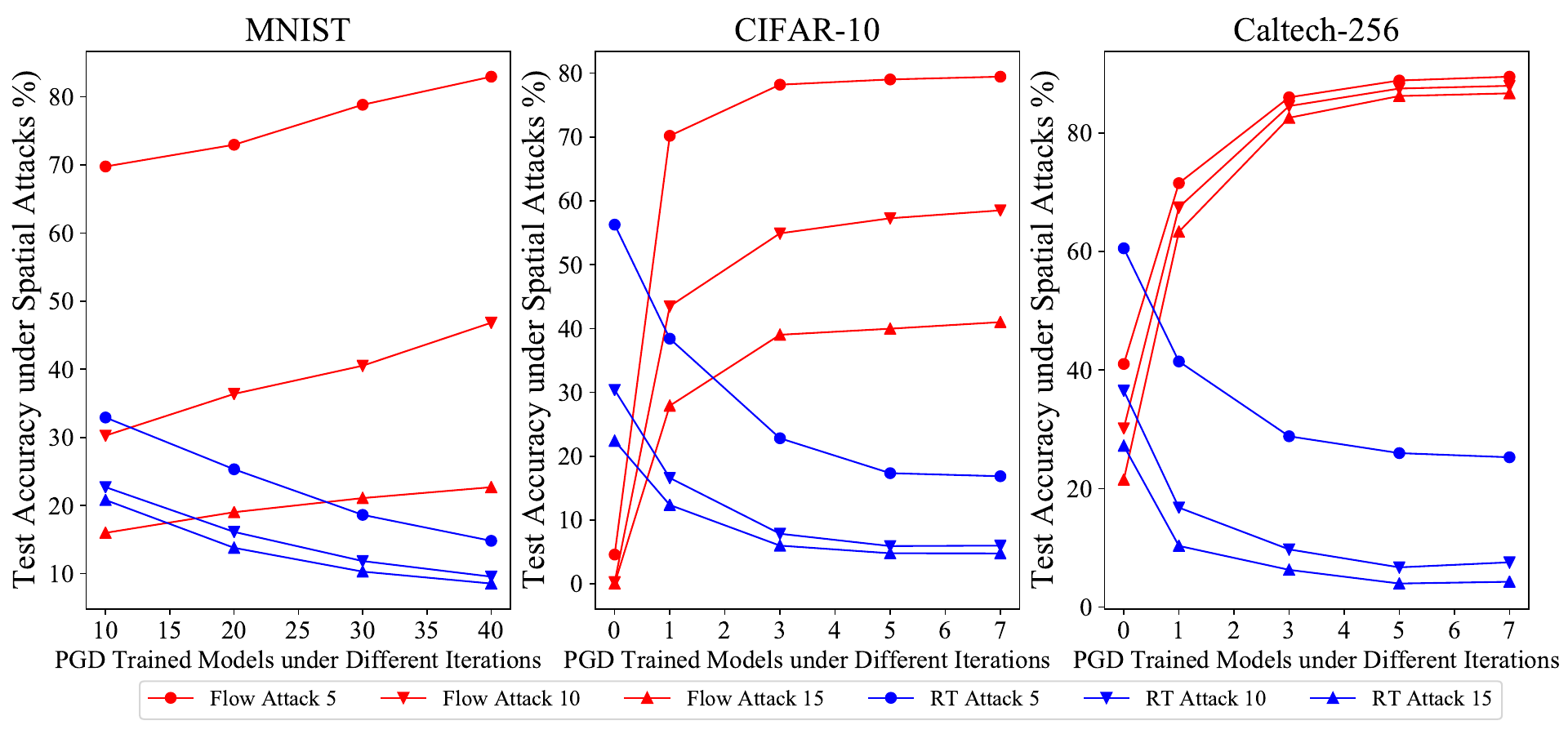}
		\caption{Relationships between sensitivity and two spatial robustness for three datasets. The X-axis represents adversarially PGD-trained models under different numbers of PGD iterations to measure the strength of sensitivity-based robustness, while the Y-axis represents the test accuracy under Flow Attack~(red) and RT Attack~(blue) with different iterations to measure the spatial robustness.}
		\label{figure_3relationship}
	\end{figure*}
	
	\section{Relationship Between Sensitivity and Spatial Robustness}
	
	In this section, we will empirically investigate the relationships between different robustnesses and then explain them from the perspective of shape-based representation by leveraging a saliency map.
	
	\subsection{Relationships}
	
	We conduct rigorous experiments on MNIST, CIFAR-$10$, and Caltech-$256$ datasets to empirically examine the behavior of local and global spatial robustness as the sensitivity-based robustness increases. Specifically, after adversarially training multiple PGD~(sensitivity-based) robust models with different numbers of PGD iterations, we further compute their test accuracy under Flow-based and RT-based spatial attacks via methods proposed in Section~\ref{sec:spatial}. The accuracy is computed on correctly classified test data for the model under consideration to mitigate the impact of the slightly different generalizations of these PGD-trained models. We fix both $\epsilon_{F}$ and $\epsilon_{RT}$  as 0.3 on MNIST, and choose $\epsilon_{F}$ and $\epsilon_{RT}$ as 0.3 and 1.0, respectively, on CIFAR-$10$ and Caltech-$256$. Then, we can control their strength of perturbations by adjusting the number of iterations in Flow-based and RT-based spatial attacks.

	In Figure~\ref{figure_3relationship}, the X-axis shows adversarially PGD-trained models with different numbers of PGD iterations, which can measure the different strengths of a model's PGD~(sensitivity-based) robustness. The Y-axis represents the computed test accuracy of the corresponding PGD-trained models under different spatial attacks, and a high-level test accuracy reflects a model's high spatial robustness. It turns out that Flow-based spatial robustness~(red lines) presents a steady ascending tendency across three datasets as the PGD sensitivity-based robustness increases, while the trend of RT-based spatial robustness~(blue lines) fluctuates conversely. This result reveals that the sensitivity-based and RT-based spatial robustness is a trade-off relationship, consistent with the previous conclusion~\cite{kamath2020invariance,tramer2019adversarial}. However, this trade-off does not ~(even on the contrary) apply to the local spatial sensitivity, where sensitivity-based and Flow-based spatial robustness is positively correlated. We provide strong and intuitive evidence from the perspective of shape-biased representation below.

	\begin{figure}[b!]
		\centering
		\includegraphics[width=0.6\textwidth,trim=100 75 100 90,clip]{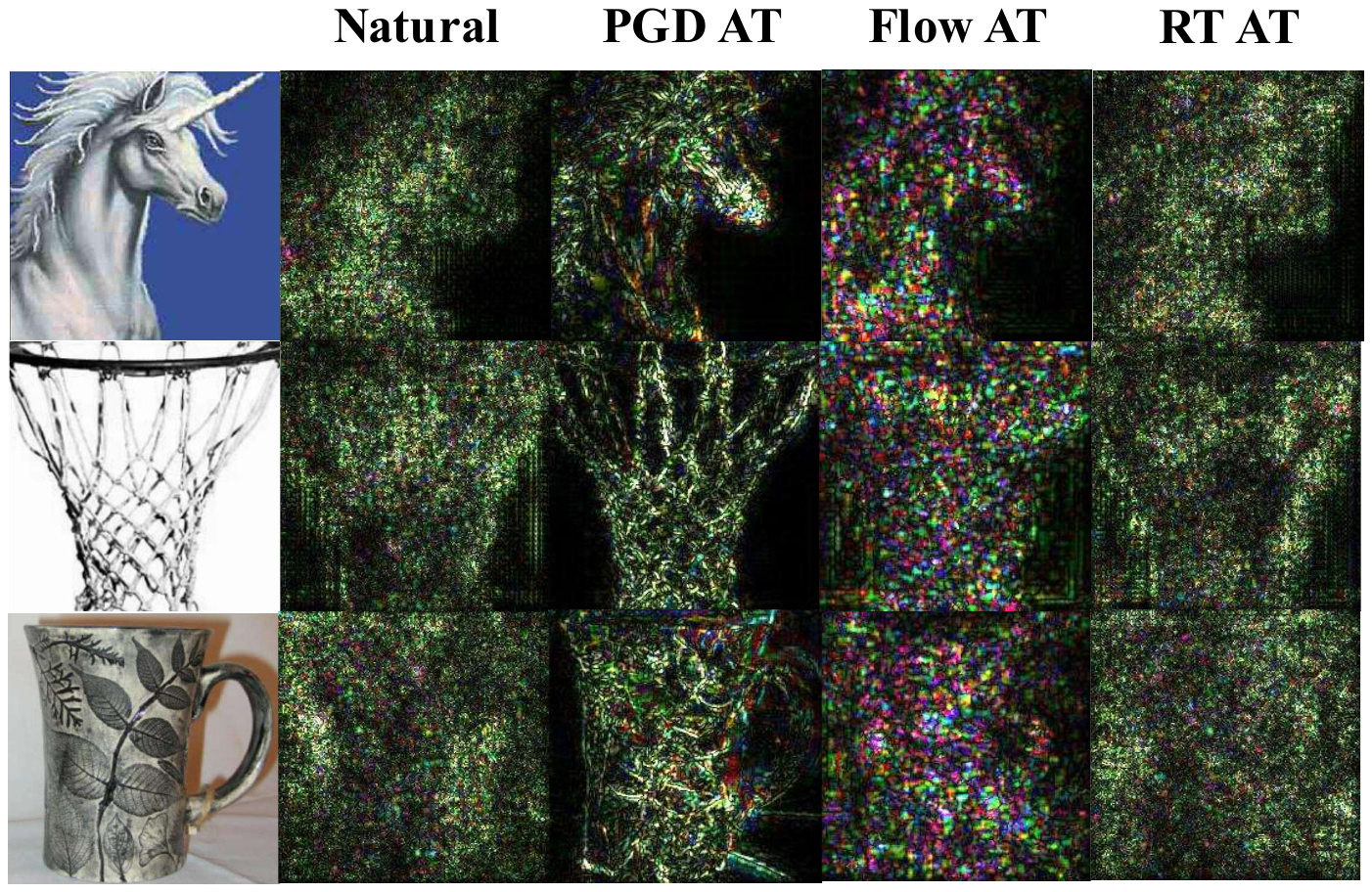}
		\caption{Saliency maps of four types of training models on some randomly selected images on Caltech-$256$.}
		\label{figure_4representation}
	\end{figure}
	
	\subsection{Explanation from the Shape-bias Representation}

	\begin{figure*}[b!]
		\centering
		\includegraphics[width=1.0\textwidth,trim=0 60 0 55,clip]{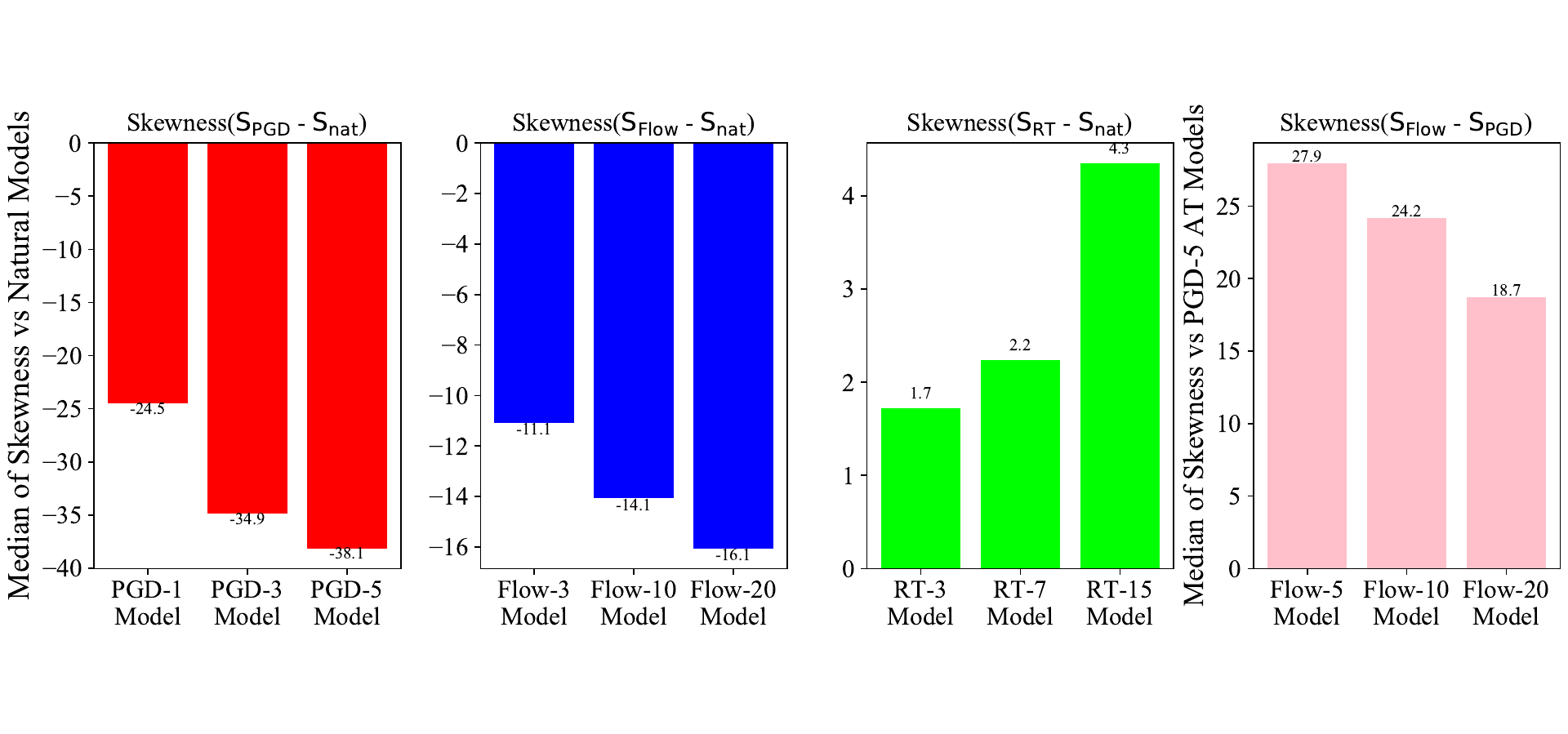}
		\caption{Median of skewness of saliency maps difference among robust models across all test data as compared with other models. The first three sub-pictures are compared with the naturally trained model, while the last one is compared with the PGD-trained model.}
		\label{figure_5skewness}
	\end{figure*}

	We show first with our brief conclusion: the sensitivity-based robustness corresponds to the sparse and shape-bias representation~\cite{shi2020informative,zhang2019interpreting}, indicating that sensitivity-based robust models rely more on the \textit{global shape} during prediction rather than the \textit{local texture}. Nevertheless, the local and global spatial robustness are associated with different representation manners. 
	
	We visualize the saliency maps of naturally trained, PGD, Flow-based, and RT adversarially trained models on some randomly selected images on Caltech-$256$, which are exhibited in Figure~\ref{figure_4representation} to examine the shape-biased representation. Specifically, visualizing the saliency maps aims at assigning a sensitivity value, sometimes also called ``attribution'', to show the sensitivity of the output to each pixel of an input image. Following \cite{shi2020informative,zhang2019interpreting}, we leverage SmoothGrad~\cite{smilkov2017smoothgrad} to calculate the saliency map $S(x)$ of an image $x$, which alleviates the noises in the gradient by averaging over the gradient of $n$ noisy copies of an input:
	\begin{equation}\label{eq_saliencemap}
		\begin{aligned} 
			S(x)=\frac{1}{n} \sum_{i=1}^{n} \frac{\partial f^y_{\theta}\left(x_{i}\right)}{\partial x_{i}},
		\end{aligned} 
	\end{equation}	
	where $x_i=x+q_i$, and $q_i$ are noises drawn i.i.d from a Gaussian distribution $\mathcal{N}(0,\sigma^2)$. In our experiment, we set $n=100$ and the noise level $\sigma/(x_{max}-x_{min})=0.1$.

	Figure~\ref{figure_4representation} shows that PGD-trained models tend to learn a \textit{sparse and shape-biased} representation for all pixels of an image, while two types of spatially adversarially trained models suggest a converse representation. In particular, the representation from the Flow-based training model presents a \textit{noisy and shape-biased} one as it places \textit{extreme values}, although noisy, on pixels around the shape of objects, e.g., the edge between the horse and the background shown in Flow AT in Figure~\ref{figure_4representation}. On the contrary, RT-based models rely less on the shape of objects, and the saliency values tend to be \textit{dense}, \textit{smoothly} scattering around more pixels of an image.

	We calculate the distance of saliency maps from different models across all test data on the Caltech-$256$ dataset and then compute their skewness in Figure~\ref{figure_5skewness}. Specifically, we compute the pixel-wise distance between the saliency maps of the two models, and then we calculate the median of the skewness of the saliency map difference for all test data. Note that if two saliency maps have no statistical difference, then the difference in the values will follow a symmetric normal distribution with skewness 0. Negative skewness indicates that the original saliency map~(representation) is sparse as compared to the model under consideration. We plot the tendency of skewness as the strength of some specific robustness increases in Figure~\ref{figure_5skewness}. We summarize the observations into two conclusions:
	
	\begin{enumerate}
		\item Based on the first and fourth sub-pictures, both PGD and Flow-based robust models tend to learn a sparse and shape-biased representation compared with the natural model. However, the Flow-based trained model is less sparse (we call it noisy shape-biased) in comparison with the PGD-trained one. 
		\item In contrast, RT-based robust models tend to learn a dense representation. This is intuitive because the RT-trained model is expected to \textit{memorize} broader pixel locations to cope with potential rotations and transformations in the test data. 
	\end{enumerate}
	
	Overall, the divergent representation~(sparse vs. dense) between RT-based and sensitivity robustness verifies that the trade-off shown in Figure~\ref{figure_3relationship} is fundamental. More importantly, the positive correlation of sensitivity-based and local spatial robustness, shown in Figure~\ref{figure_3relationship}, can also be explained by their similar shape-biased representation, although the latter tends to be noisy.

	\section{Pareto Adversarial Robustness}
	
	\subsection{Motivation}

	\noindent \textbf{Multi-objective Optimization}.  Given the insights garnered from our analysis of the relationships between natural accuracy and different kinds of adversarial robustness, a natural question that comes up is how to design a training strategy that can perfectly balance their mutual impacts, which mainly results from their different representation manners. In most cases, their relationships exhibit trade-offs, except for the positive correlation between sensitivity robustness and local spatial robustness. We use $\mathcal{L}_{nat},\mathcal{L}_{\text{PGD}},\mathcal{L}_{\text{Flow}}$ and $\mathcal{L}_{\text{RT}}$ to represent the natural loss, the PGD adversarial loss, the Flow-based and the RT-based adversarial loss, respectively.  We cast obtaining universal adversarial robustness as well as maintaining natural generalization ability as a multi-objective optimization problem~\cite{leung2018collaborative}, encompassing all of the aforementioned losses with a loss vector:
	\begin{equation}\label{eq_multioptimization}
		\begin{aligned} 
			\min_\theta \mathcal{L}^\theta = (\mathcal{L}_0^\theta,\mathcal{L}_1^\theta,\mathcal{L}_2^\theta,\mathcal{L}_3^\theta)^\top,
		\end{aligned} 
	\end{equation}	
	where $\mathcal{L}_0^\theta,\mathcal{L}_1^\theta,\mathcal{L}_2^\theta,\mathcal{L}_3^\theta$ represent $\mathcal{L}_{\text{nat}},\mathcal{L}_{\text{PGD}},\mathcal{L}_{\text{Flow}},\mathcal{L}_{\text{RT}}$ respectively for simplicity, sharing the same model parameter $\theta$. The multi-objective optimization is to optimize all loss functions simultaneously by exploiting the shared knowledge and structure, e.g., the representation.

	\noindent \textbf{Pareto Optimization}. To harmonize these competing optimization objectives in the context of adversarial robustness, we introduce Pareto Optimization~\cite{kim2005adaptive,lin2019pareto,li2014pareto}, which is successfully applied when optimal decisions need to be taken in the presence of trade-offs between multiple conflicting objectives. Pareto optimization endeavors to achieve Pareto Optimality, a balanced situation between all objectives, \textit{where none of the objective functions can be improved in value without degrading some of the other objective values}. Mathematically, we have the following definitions~\cite{zitzler1999multiobjective,lin2019pareto}.
	
	\noindent \textbf{Pareto Dominance in Adversarial Robustness}. Let $\theta^1, \theta^2$ be two parameters in the space $\Omega$. $\theta^1$ dominates $\theta^2$, i.e., $\theta^1 \prec \theta^2$, if and only if $\mathcal{L}_i^{\theta^1} \leq \mathcal{L}_i^{\theta^2}, \forall i \in \{0, 1, 2, 3\}$ and $\mathcal{L}_j^{\theta^1} < \mathcal{L}_j^{\theta^2}, \exists j \in\{0, 1, 2, 3\}$.
	
	\noindent \textbf{Pareto Optimality}. $\theta^*$ is a Pareto optimal point, and $\mathcal{L}^{\theta^*}$ is a Pareto optimal objective vector if it does not exist $\hat{\theta} \in \Omega$ such that $\hat{\theta} \prec \theta^*$. The resulting \textit{ Pareto front} contains all Pareto optimal solutions.
	

	\noindent \textbf{Pareto Adversarial Robustness.} Based on the insights presented above, a natural approach for incorporating Pareto criteria into multi-objective optimization in the context of adversarial training is to achieve universal adversarial robustness as well as maintain a desirable natural accuracy. The resulting Pareto Front contains all optimal, adversarially trained models for the given different constraints. The detailed formulation is presented later in Section~\ref{sec:paretoat}.
	
	\begin{figure}[b!]
		\centering 
		\includegraphics[width=0.45\textwidth,trim=0 0 40 25,clip]{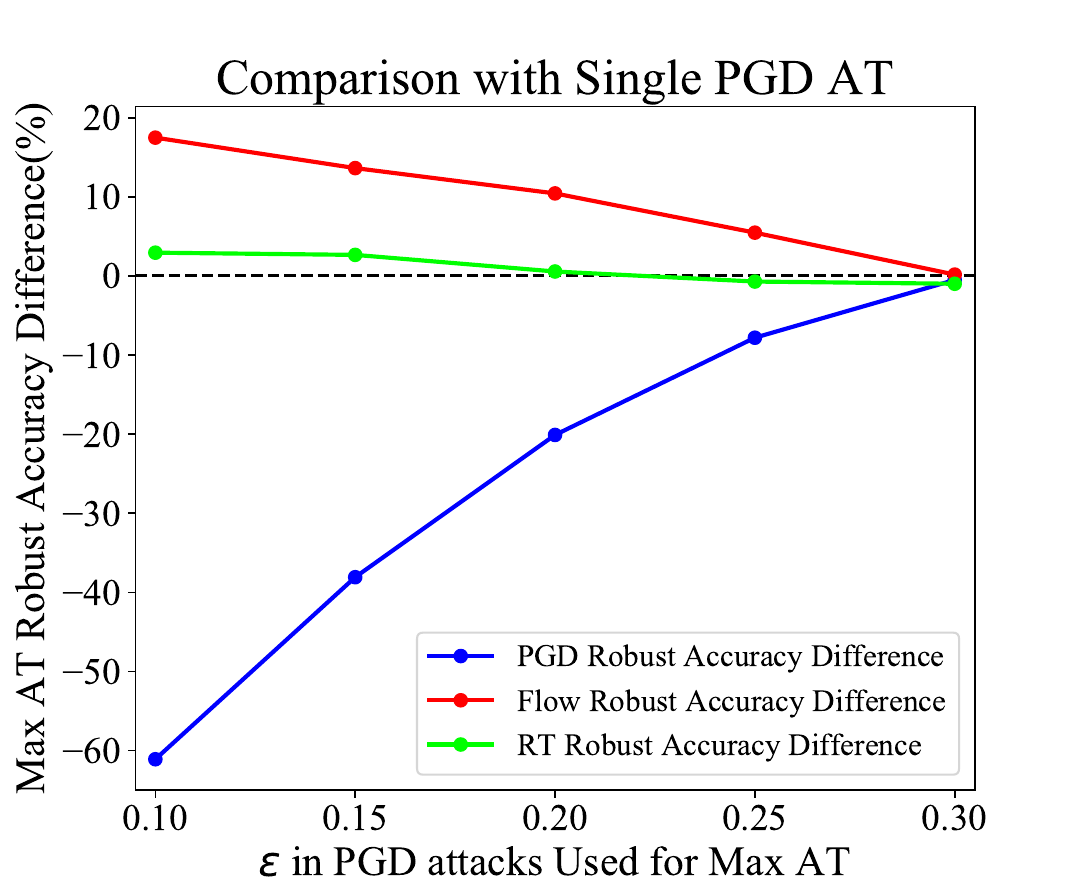}
		\caption{The difference between the model trained by the PGD method and Max AT with different parameter $\epsilon$ for the PGD attack in the PGD adversarial training.}
		\label{figure_6max}
	\end{figure}
	
	\subsection{Limitations of the Existing Strategies.} 
	
	We denote $\mathcal{R}_{\text{adv}}(f ; S_i):=\mathbb{E}_{(x, y) \sim \mathcal{D}}\left[\max _{\boldsymbol{r} \in S_i} \mathcal{L}(f(x+r), y)\right]$ as the adversarial risk under perturbation sets $S_i, i=1,...,m.$ Our goal is to find $f_{\theta}$ that can achieve uniform risk minimization across all $S_i$ as well as the minimal risk in the natural data. There are two common strategies to handle this issue.
	
	\noindent \textbf{1) Average adversarial training~(Ave AT)}~\cite{tramer2019adversarial}. $\mathcal{R}_{\text{ave}}(f ; S):=\mathbb{E}_{(x, y) \sim \mathcal{D}}\left[\frac{1}{m}\sum_{i=1}^{m} \max _{\boldsymbol{r} \in S_i} \mathcal{L}(f(x+r), y)\right]$, regards each adversarial robustness as having equal status. Intuitively, it may yield unsatisfactory solutions when the strength of different attacks mixed in the training are not balanced. 
	
	\noindent \textbf{2) Max adversarial training~(Max AT)} ~\cite{tramer2019adversarial,maini2019adversarial}, i.e., $\mathcal{R}_{\text{max}}(f ; S):=\mathbb{E}_{(x, y) \sim \mathcal{D}}[\max_{i} \{ \max _{\boldsymbol{r} \in S_i} \mathcal{L}(f(x+r), y)\}]$ tries to optimize over the max loss from the largest perturbations.

	\noindent \textbf{Overfitting issue of Max AT}. Intuitively, Max AT may overfit to one specific type of adversarial robustness if its adversarial attack used for training is too strong. In Figure~\ref{figure_6max}, we plot the difference in robust accuracy between Max AT and single PGD adversarial training. It turns out that as the strength of PGD attack $\epsilon$ used in Max AT increases, the difference among the three kinds of robust accuracy between Max AT and a single PGD AT tends to vanish. This indicates that the comprehensive robustness of Max AT degenerates to a single PGD adversarial training because the PGD loss tends to dominate as the strength of the PGD attack increases.
	

	\noindent \textbf{Overfitting issue of Ave AT based on its Relationship with Max AT.} We consider the generalization issue based on different risks and then set the risk in Max AT and Ave AT as $\mathcal{R}_{\text{max}}=\max_{i} \mathcal{R} (f,S_i)=\max_{i} \mathcal{R}^{S_i}$ and $\mathcal{R}_{\text{ave}}=\frac{1}{m} \sum_{i=1}^{m}\mathcal{R}(f_{\theta},S_i)=\frac{1}{m} \sum_{i=1}^{m} \mathcal{R}^{S_i}$. Proposition~\ref{prop:affine} informs that Max AT is closely associated with some form of Ave AT. This indicates that Max AT is likely to perform similarly to the specific form of Ave AT, which also suffers from unsatisfactory solutions when the strength of different attacks mixed in training is imbalanced. We verified this claim in Table~\ref{table_score2} under a stronger PGD attack in Section~\ref{sec:paretoat}.
	
	\begin{prop}\label{prop:affine}
		Given KKT differentiability and qualification conditions, $\exists \lambda_i \ge 0$, such that the risk minimizer in Max AT, i.e., $\mathcal{R}^{\star}_{\text{max}}$ is a first-order stationary point of $\sum_{S_{i} \in \mathcal{S}} \lambda_{i} \mathcal{R}^{S_i}$ regardless of the relationship of $S_{i}$.
		\label{prop:maxAT}
	\end{prop}
	
	\noindent \textbf{Remark.} We point out that both Ave AT and Max AT may suffer from the robustness overfitting issue and thus fail in certain scenarios. However, a clever combination choice among all involved adversarial losses has the potential to alleviate the overfitting issues, thus outperforming both Max AT and Ave AT in terms of universal robustness. Motivated by this, we propose Pareto Adversarial Training in the next section, which will provide strong empirical evidence to support this intuition.

	\subsection{Pareto Adversarial Training}~\label{sec:paretoat}
	
	We apply linear scalarization to solve the multi-objective optimization, which is the most commonly used approach. We denote $\alpha=(\alpha_0, \alpha_1, \alpha_2,\alpha_3)$ as the combination coefficients for various losses. Thus, the objective function is $\min_{\theta}   \sum_{i=0}^{3}  \mathbb{E}_x    \left[\alpha_{i}^*(\theta) \mathcal{L}_i^\theta\right]$. Further, within the context of Pareto optimality, our goal is to find \textit{optimal combinations} $\alpha$ between natural accuracy, sensitivity-based, and spatial robustness to perfectly balance their mutual impacts during the whole training process.  Furthermore, we train a model $f_{\theta}$ under the optimal combinations $\alpha^*$ of different losses, and the computation of $\alpha^*$ in training is also associated with different losses determined by model parameters $\theta$. This implies a bilevel optimization problem with $\theta$ as the upper-level variable and $\alpha$ as the lower-level variable. In the construction of low-level optimization regarding $\alpha$, we apply a two-moment objective function concerning all losses. We name this bi-level optimization as \textit{Pareto Adversarial Training}, which is formulated as:
	\begin{equation}
		\begin{aligned}\label{eq_alpha}
			&\min_{\theta}   \sum_{i=0}^{3}  \mathbb{E}_x    \left[\alpha_{i}^*(\theta) \mathcal{L}_i^\theta\right], \\ &\ \text{s.t.}  \ \alpha^*= \argmin_{\alpha} \sum_{i=0}^{3}\sum_{j=0}^{3}\mathbb{E}_x \left[(\alpha_{i}\mathcal{L}_i^\theta - \alpha_{j} \mathcal{L}_j^\theta)^2\right], r =  \sum_{i=1}^{3}\alpha_{i} \mathbb{E}_x \left[\mathcal{L}_i^\theta\right], \sum_{i=0}^{3} \alpha_{i}=1, \alpha_{i} \geq 0, \forall i=0,1,2,3,
		\end{aligned}
	\end{equation}
	where $r$ indicates the expectation of one-moment over \textit{all robust losses, i.e., spatial and sensitivity-based losses}, which reflects the strength of comprehensive robustness we require after solving this quadratic lower-level optimization regarding $\alpha$. In particular, given the model parameter $\theta$ in each training step, the larger $r$ we require will push the resulting $\alpha_{i}, i=1,2,3$ larger, thus increasing the weight of the robust losses rather than the natural loss to pursue more robustness. 
	
	\begin{algorithm}[t!]
		\caption{Bi-level Optimization in Pareto Adversarial Training.}
		\textbf{Input}: Training data ($\mathcal{X}$, $\mathcal{Y}$). Batch size $M$ and adjustable hyper-parameter $r$. Initialization of $\alpha$ as $[1/4, 1/4, 1/4, 1/4]$. \\
		\textbf{Output}: Classifier $f_\theta$. 
		\begin{algorithmic}[1] 
			\REPEAT
			\setstretch{0.8}
			\STATE Sample $\{\mathbf{y}_1,...,\mathbf{y}_M \}$ and $\{\mathbf{x}_1,...,\mathbf{x}_M \}$ from all training data.
			
			\STATE \textbf{/ * Step 1: Compute loss in Eq.~\ref{eq_alpha} * /}
			
			\STATE Compute natural loss $\mathcal{L}_{\text{nat}}$, and adversarial loss $\mathcal{L}_{\text{PGD}},\mathcal{L}_{\text{Flow}},\mathcal{L}_{\text{RT}}$ based on natural cross entropy loss, PGD loss and Eq.~\ref{eq_flow_ce} and \ref{eq_RT_optimization}, respectively.
			
			\STATE  \textbf{/ * Step 2: Upper-level Optimization over $\theta$ * /}	
			
			\STATE Given the current $\alpha$, update $f_\theta$ by descending its stochastic gradient of:		
			\begin{equation}
				\begin{aligned}
					\frac{1}{M} \sum_{i=1}^{M} \mathcal{L}^{\text{CE}}(f_{\theta}(\mathbf{x}_i), \mathbf{y}_i) & = \frac{1}{M} \sum_{i=1}^{M} \alpha_0 \mathcal{L}_{\text{nat}}(f_{\theta}(\mathbf{x}_i), \mathbf{y}_i) + \alpha_1 \mathcal{L}_{\text{PGD}}(f_{\theta}(\mathbf{x}_i), \mathbf{y}_i) \\
					&+ \alpha_2 \mathcal{L}_{\text{Flow}}(f_{\theta}(\mathbf{x}_i), \mathbf{y}_i) + \alpha_3 \mathcal{L}_{\text{RT}}(f_{\theta}(\mathbf{x}_i), \mathbf{y}_i)
				\end{aligned}\nonumber
			\end{equation}	
			
			\STATE  \textbf{/ * Step 3: Lower-level Optimization over $\alpha$ * /}
			
			\STATE Compute $\hat{\mu}$ and $\hat{\Sigma}$ by sliding window technique in Eq.~\ref{eq_alpha_appendix}.
			
			\STATE Evaluate $P$ in the quadratic form shown in Eq.~\ref{eq_alpha_appendix}.
			
			\STATE Solve Eq.~\ref{eq_alpha_appendix} via CVXOPT tool to obtain the $\alpha$.		
			
			\UNTIL Convergence		
		\end{algorithmic}
		\label{alg:framework}
	\end{algorithm}
	
	\noindent \textbf{Two-Moment Objective Function.} The two-moment form is a common practice in Pareto optimization. For example, in the financial portfolio theory, the mean-variance optimization is normally leveraged to compute the \textit{Pareto Efficient Front}, where the risk of the asset portfolio, measured by their variances, is minimized to balance the different correlations of these assets given an expected return from the investor. Similarly, the square loss of the difference between each loss pair in Eq.~\ref{eq_alpha} measures their mutual impacts. For instance, a decrease in $\mathcal{L}_{\text{PGD}}$ tends to increase $\mathcal{L}_{\text{RT}}$ as they have a fundamental trade-off relationship. We hope to mitigate all these mutual impacts, measured by the weighted quadratic differences, among all losses given an expected robustness level of $r$. In the implementation, as we regard all losses as random variables with their stochasticity arising from the mini-batch sampling from data, we leverage the sliding windows technique to compute their expectations. Our bi-level optimization within a batch is (1) $\theta$: update parameters $\theta$ via SGD and (2) $\alpha$: solve $\alpha$ via quadratic programming. Denote the random variables $\mathcal{L}_0,\mathcal{L}_1,\mathcal{L}_2,\mathcal{L}_3$ with mean vector $\mu$ and covariance matrix $\Sigma$. We transform our lower-level optimization regarding $\alpha$ as the following standard quadratic form:
	\begin{equation}
		\begin{aligned}\label{eq_alpha_appendix}
			\min_{\theta, \alpha} & \ \ \alpha^T P \alpha \quad 
			\text { s.t. } &
			\left[\begin{array}{cccc}
				0 & \mu_1 & \mu_2 & \mu_3\\
				1 & 1 & 1 & 1\\
			\end{array}\right] \alpha = 
			\left[\begin{array}{c}
				r\\
				1
			\end{array}\right],  -\alpha \leq 
			\textbf{0},
		\end{aligned}
	\end{equation}
	where $P=8(\text{diag}(\Sigma)+\text{diag}(\mu \mu^T))-2(\Sigma+\mu \mu^T)$. We utilize the $\textit{CVXOPT}$ tool to solve this quadratic optimization within each mini-batch training.  $\textit{CVXOPT}$ is probably the most popular free software package for convex optimization based on the Python programming language that can solve quadratic programming effectively. We also provide proof of the quadratic formulation in \ref{appendix:pareto_optimization}.

	A detailed algorithm description is given in Algorithm~\ref{alg:framework}. In the lower-level procedure of Pareto adversarial training, we solve the quadratic optimization regarding $\alpha$ given $\theta$ in each training step to obtain the optimal combinations among natural loss, sensitivity-based, and spatial adversarial loss. Then in the upper-level optimization, we leverage our familiar SGD method to update $\theta$ based on $\alpha^*$ calculated from the lower-level problem. Note that the computation complexity of our method is similar to Ave AT and Max AT, which are still competitive in computation.
	
	\subsection{Approximated Pareto Front}\label{experiment:paretofront}
	
	\begin{table*}[htbp]
		\centering
		\scalebox{0.7}{
			\begin{tabular}{cccccccc}
				\toprule[1pt]
				\textbf{Dataset} & \textbf{Robustness Score~($\%$)}  & \textbf{Natural Model} & \textbf{PGD AT} &  \textbf{Spatial AT} & \textbf{Max AT}&\textbf{Ave AT}&\textbf{Pareto AT}~($r=2.2$)\\
				\hline
				\multirow{4}*{MNIST}&Sensitivity-based Robustness&29.40&98.42&0.24&65.16&92.70&88.06\\
				~&Local Spatial Robustness&14.36&38.23&27.59&53.02&48.51&58.70\\
				~&Global Spatial Robustness&16.70&12.77&78.76&51.47&88.76&90.40\\
				~&\textbf{Universal Robustness}&0.0&88.97&46.14&109.19&169.52&\textbf{176.71}\\
				\hline
				\textbf{Dataset} & \textbf{Robustness Score~($\%$)}  & \textbf{Natural Model} & \textbf{PGD AT} &  \textbf{Spatial AT} & \textbf{Max AT}&\textbf{Ave AT}&\textbf{Pareto AT}~($r=4.0$)\\
				\hline
				\multirow{4}*{CIFAR-10}&Sensitivity-based Robustness&0.82&70.24&12.11&52.04&49.68&51.65\\
				~&Local Spatial Robustness&9.34&72.29&83.63&85.31&79.88&80.38\\
				~&Global Spatial Robustness&57.28&18.94&40.96&40.06&64.98&66.36\\
				~&\textbf{Universal Robustness}&0.0&94.04&69.26&109.98&127.11&\textbf{130.96}\\
				\hline
				\textbf{Dataset} & \textbf{Robustness Score~($\%$)}  & \textbf{Natural Model} & \textbf{PGD AT} &  \textbf{Spatial AT} & \textbf{Max AT}&\textbf{Ave AT}&\textbf{Pareto AT}~($r=6.0$)\\
				\hline
				\multirow{4}*{Caltech-256}&Sensitivity-based Robustness&4.74&82.43&6.94&59.81&71.60&76.52\\
				~&Local Spatial Robustness&34.59&87.96&88.75&65.89&86.67&87.39\\
				~&Global Spatial Robustness&49.73&21.71&65.04&64.64&53.68&50.00\\
				~&\textbf{Universal Robustness}&0.0&103.05&71.66&101.28&122.89&\textbf{124.85}\\
				\bottomrule[1pt]
			\end{tabular}
		}
		\caption{Robustness Score of each type of adversarial robustness on MNIST, CIFAR-10, and Caltech-256. Each type of robustness~($\%$) is the average test accuracy under different strengths of perturbations. We choose the universal robustness of the Natural Model as the baseline and set it as 0. We use the difference in average test accuracy between other models and the Natural Model and then sum them as the universal robustness.}
		\label{table_score}
	\end{table*}
	
	\begin{figure*}[b!]
		\centering
		\includegraphics[width=0.9\textwidth,trim=10 50 10 20,clip]{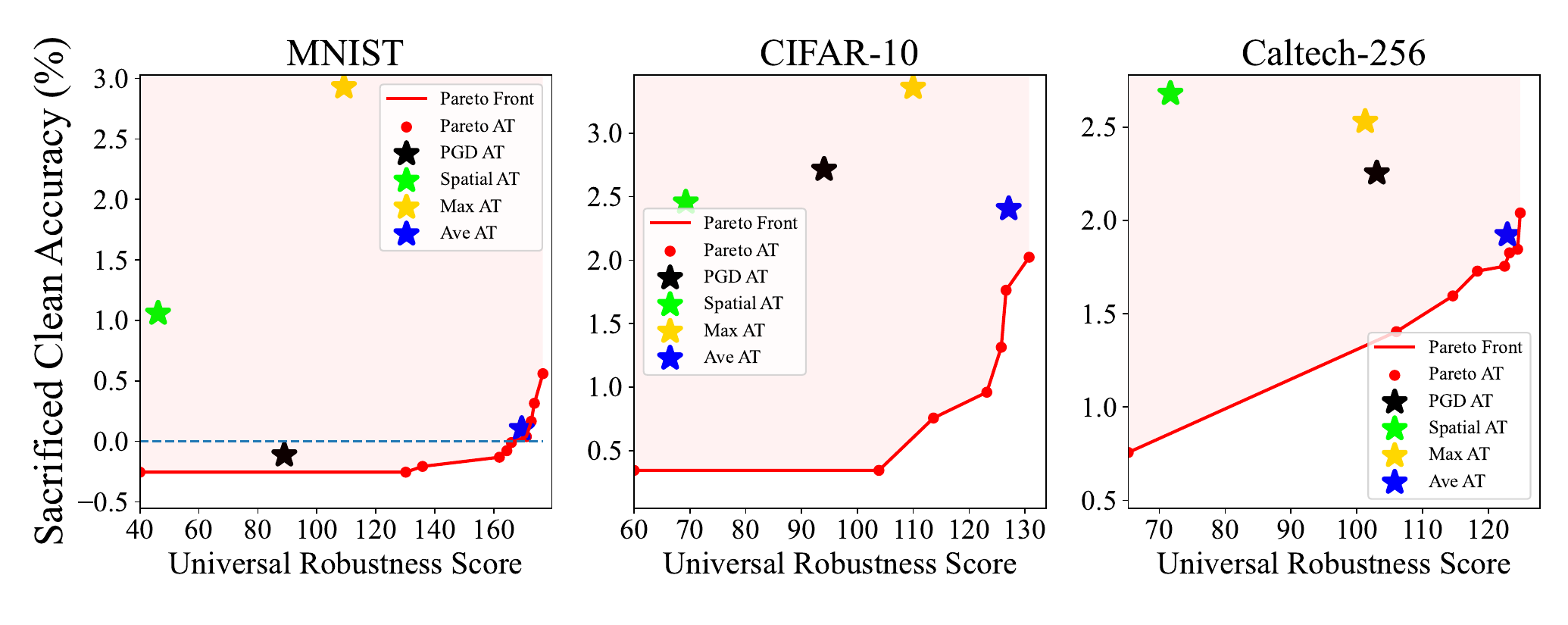}
		\caption{The Pareto front~(red lines) between the universal robustness score and sacrificed clean accuracy on MNIST, CIFAR-$10$, and Caltech-$256$. The vertical axis is the decrease of the natural accuracy compared with the naturally trained model and has been under the log transformation along two directions.}
		\label{figure_7front}
	\end{figure*}

	By adjusting the upper bound of the expected adversarial robustness loss $r$, we can evenly generate Pareto optimal solutions where the obtained models will have different levels of robustness under optimal combinations. The set of all Pareto optimal solutions then forms the \textit{Pareto front}. Rigorously, it is almost impossible to attain all Pareto optimal solutions for a general continuous multi-objective optimization problem unless a closed-form solution exists for each $r$. Alternatively, we leverage the limited solutions obtained by solving a series of multi-objective optimization problems for various $r$ to approximate the Pareto front.

	Thus, we train deep neural networks under different adversarial training strategies, i.e., PGD Adversarial Training~(PGD AT), Spatial Adversarial Training~(Spatial AT) proposed in Section~\ref{spatialAT}, Max AT, Ave AT, and Pareto Adversarial Training~(Pareto AT) under different $r$, in which we apply a proper iteration. Then we evaluate their test accuracy under PGD, Flow-based, and RT attacks under different perturbation strengths. Next, we average the test accuracies for each type of attack, and the result is a quantitative measure of the specific robustness, called \textit{Robustness Score}. To evaluate the universal robustness, we further compute the average of Robustness Scores for all kinds of robustness and use the increment over the naturally trained model as the metric called \textit{Universal Robustness Score}. We report the Robustness Scores of all models on CIFAR-10 in Table~\ref{table_score}, and the results on the other two datasets are similar. All implementation details are provided in \ref{appendix:implementation}. It shows that Pareto AT~($r=4.0$) has the best universal robustness score among all the models considered, although the highest specific robustness normally exists in the adversarial training model that only focuses on it.

	Finally, we plot the universal robustness scores and the sacrificed clean accuracy of all methods across three datasets in Figure~\ref{figure_7front}, where multiple Pareto AT models~(red points) are trained under different $r$. The Pareto criterion exhibited in Figure~\ref{figure_7front} provides an optimality principle, which enables Pareto Adversarial Training to achieve the best universal robustness among all the methods considered, given a certain tolerable level of sacrificed clean accuracy. By adjusting the expected universal robustness $r$ in Pareto Adversarial Training, we can develop the set of Pareto optimal solutions, i.e., the Pareto front.  It shows that all other methods are above our Pareto front and are less effective than our proposal.

	\begin{table*}[htbp]
		\centering
		\scalebox{0.8}{
			
			\begin{tabular}{cccccccc}
				\toprule[1pt]
				\textbf{Robustness Score~($\%$)}  & \textbf{Natural Model} &\textbf{Ave AT}& \makecell{\textbf{Pareto AT}\\($r=3.5$)}&\makecell{\textbf{Pareto AT}\\($r=3.7$)}& \makecell{\textbf{Pareto AT}\\($r=4.0$)}& \makecell{\textbf{Pareto AT}\\($r=4.2$)}\\
				\hline
				Natural Accuracy &91.43&56.39&82.64&79.69&71.68&61.53\\
				\hline
				Sensitivity-based &0.82&64.11&53.73&58.70&63.28&65.19\\
				Local Spatial &9.34&82.45&80.10&77.62&81.01&82.38\\
				Global Spatial &57.28&51.36&66.57&67.56&59.69&52.04\\
				\hline
				Universal Robustness&0.0&197.92&200.39(+2.37)&203.88(+5.96)&\textbf{203.98}(+6.06)&199.61(+1.69)\\
				\bottomrule[1pt]
			\end{tabular}
		}
		\caption{Robustness Score on CIFAR-10 with a \textbf{larger} step size 8/255 and $\epsilon$ as 16/255 in PGD perturbations used for both Ave AT and Pareto AT across different $r$.}
		\label{table_score2}
	\end{table*}
	
	\noindent \textbf{Overfitting Issue of Ave AT.} Note that although the perturbation strength adopted in Table~\ref{table_score} is mild, we need to point out that \textit{the superiority of Pareto AT over Ave AT can be higher if the overfitting issue is severe}. We demonstrate this claim in Table~\ref{table_score2}, where we apply a stronger PGD perturbation used in AT. Finally, we find that Ave AT overfits sensitivity robustness more severely, achieving much less universal robustness and sacrificing more clean accuracy than Pareto AT. Pareto Adversarial Training can mitigate the overfitting issue regarding an overly strong perturbation in AT because Pareto AT can automatically adjust the weights $\alpha$ while training, which is the key advantage of Pareto AT over Ave AT. 
	
	\noindent \textbf{Sensitivity Analysis.} Comparing universal adversarial robustness between Table~\ref{table_score} and Table~\ref{table_score2}, it can be seen that Pareto AT achieves more consistent universal adversarial robustness. In addition to this sensitivity analysis in terms of perturbation sizes, we also investigate the variation of universal adversarial robustness by changing the expected adversarial robustness loss $r$. Results are provided in Table~\ref{table_score2}. It suggests that Pareto AT with a mild $r$ can achieve the best universal robustness score, while Pareto AT with an excessively large or small $r$ may not have sufficient universal robustness. Moreover, Pareto Front in Figure~\ref{figure_7front} also serves as the sensitivity analysis results in terms of different $r$.

	Overall, we conclude that Pareto adversarial training perfectly balances the mutual impacts of sensitivity-based robustness and spatial robustness under the Pareto criterion.
	
	\section{Discussion and Conclusion}
	The principal purpose of our work is to design a novel approach to achieve universal adversarial robustness. We first analyze the two main branches of spatial robustness and then integrate them into one attack and adversarial training design. Furthermore, we investigate the comprehensive relationships between sensitivity-based and two distinct spatial robustnesses from the perspective of representation. Based on the understanding of the mutual impacts of different kinds of adversarial robustness, we introduce the Pareto criterion into the adversarial training framework to develop Pareto Adversarial Training. The resulting Pareto front provides optimal solutions over existing baselines, given the universal robustness level we hope to attain. In the future, we hope to apply Pareto analysis to more general Out-of-Distribution generalization settings.

	\Acknowledgements{Z. Lin was supported by National Key R$\&$D Program of China (2022ZD0160300), the NSF China (No. 62276004), and Qualcomm.}
	%
	%
	%
	%
	%
	\bibliographystyle{plain} 
	\bibliography{Pareto}


	\clearpage
	\begin{appendix}
		\section{Visualization of Various Attacks}~\label{appendix:moreimages}
\begin{figure*}[htbp]
			\centering
			\includegraphics[width=0.8\textwidth,trim=0 120 0 90,clip]{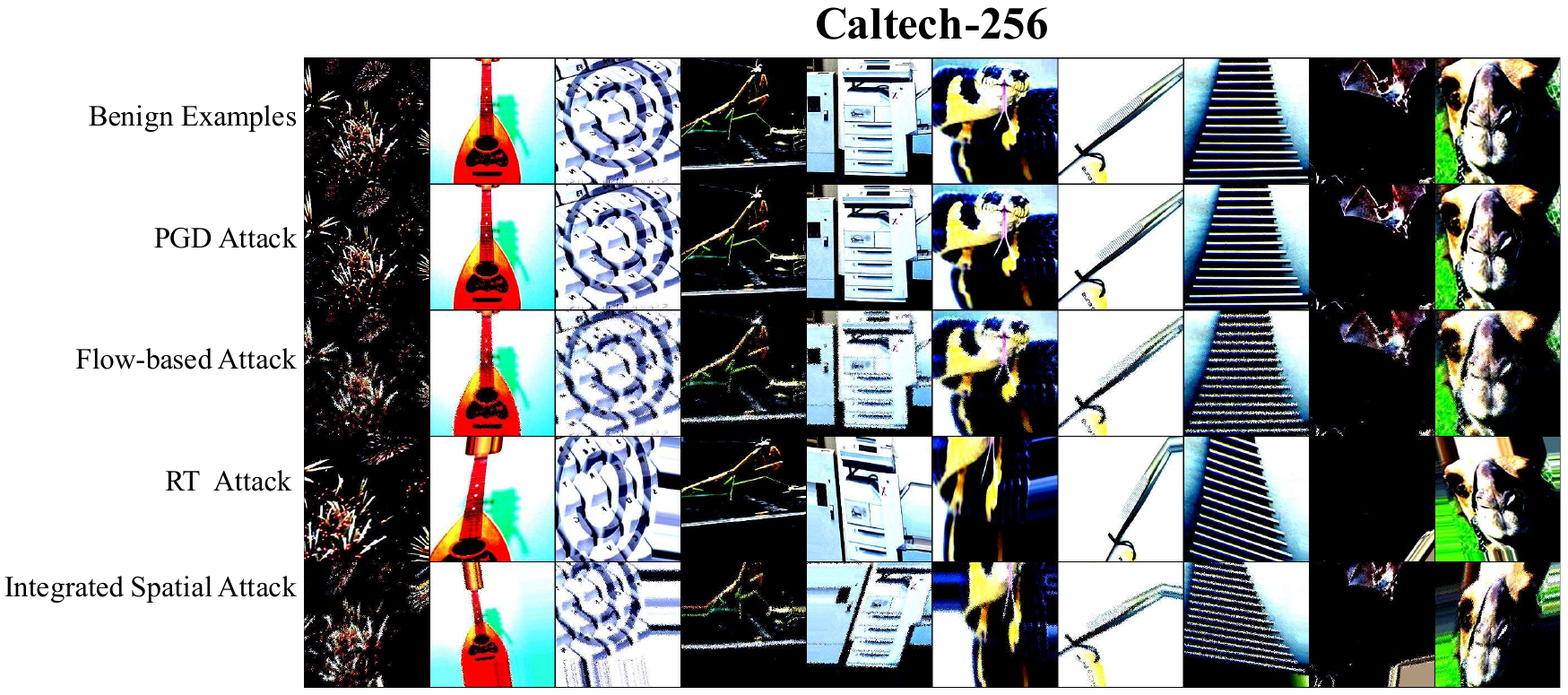}
			\caption{High-resolution images on Caltech-$256$.}
			\label{figure_9Appendix_moreimages}
		\end{figure*}

		To better present the visual effect of various kinds of adversarial attacks, we provide high-resolution results on Caltech-$256$ in Figure~\ref{figure_9Appendix_moreimages}. It turns out that Flow-based attacks focus on local spatial vulnerability that mainly blurs pixels in some local regions, while RT attacks cause a shape-based global spatial transformation. More importantly, our integrated spatial attacks are more comprehensive in the sense of spatial robustness, combining both local and local spatial sensitivity.
		
		\section{Proof of Proposition~\ref{prop:flow}}~\label{appendix:proof_flow}
		
		\begin{proof}
			Firstly, we have the following equations according to the definitions of the loss function:
			\begin{equation}\begin{aligned} 
					\mathcal{L}^{\text{CE}}_{\theta}(x_{w_F}, y)&=\log \sum_{i = 1}^{K} \exp \left(f_{\theta}^{i}(x_{w_F})\right)-f_{\theta}^{y}(x_{w_F}) \\
					\mathcal{L}^{S}_{\theta}(x_{w_F}, y)&=\log \sum_{i \neq y} \exp \left(f_{\theta}^{i}(x_{w_F})\right)-f_{\theta}^{y}(x_{w_F}) 
			\end{aligned} \end{equation}
			Then, we compute their gradients for the flow vector $x_{w_F}$. The gradient of $\mathcal{L}^{\text{CE}}_{\theta}(x_{w_F}, y)$ is shown as follows:
			\begin{equation}\begin{aligned} 
					&\nabla_{w_F} \mathcal{L}^{\text{CE}}_{\theta}(x_{w_F}, y)\\
					&=\frac{\sum_{i=1}^{K}\exp(f_{\theta}^i(x_{w_F})) \cdot \nabla_{x_{w_F}}f_{\theta}^i(x_{w_F}) \cdot \nabla_{w_F} x_{w_F}}{\sum_{i=1}^{K}\exp(f_{\theta}^i(x_{w_F}))} -  \nabla_{x_{w_F}}f_{\theta}^y(x_{w_F}) \cdot \nabla_{w_F} x_{w_F}\\
					&=\frac{\sum_{i=1}^{K}\exp(f_{\theta}^i(x_{w_F}))\nabla_{w_F} x_{w_F} (\nabla_{x_{w_F}}f_{\theta}^i(x_{w_F}) - \nabla_{x_{w_F}}f_{\theta}^y(x_{w_F}))}{\sum_{i=1}^{K}\exp(f_{\theta}^i(x_{w_F}))}
			\end{aligned} \end{equation}
			Similarly, the gradient of $\mathcal{L}^{S}_{\theta}(x_{w_F}, y)$ is:
			\begin{equation}\begin{aligned} 
					\nabla_{w_F} \mathcal{L}^{S}_{\theta}(x_{w_F}, y)=\frac{1}{\sum_{i \neq y}^{K}\exp(f_{\theta}^i(x_{w_F}))} \cdot (\sum_{i \neq y}^{K}\exp(f_{\theta}^i(x_{w_F})) 	\nabla_{w_F} x_{w_F} (\nabla_{x_{w_F}}f_{\theta}^i(x_{w_F}) - \nabla_{x_{w_F}}f_{\theta}^y(x_{w_F})))
			\end{aligned} \end{equation}
			Then we take the multiplication of $\nabla_{w_F} \mathcal{L}^{S}_{\theta}(x_{w_F}, y)$ by a term $\frac{\sum_{i \neq y}^{K}\exp(f_{\theta}^i(x_{w_F}))}{\sum_{i=1}^{K}\exp(f_{\theta}^i(x_{w_F}))}$, finally we attain:
			\begin{equation}\begin{aligned} 
					\nabla_{w_F} \mathcal{L}^{S}_{\theta}(x_{w_F}, y) \cdot \frac{\sum_{i \neq y}^{K}\exp(f_{\theta}^i(x_{w_F}))}{\sum_{i=1}^{K}\exp(f_{\theta}^i(x_{w_F}))}
					=&\frac{\sum_{i \neq y}^{K}\exp(f_{\theta}^i(x_{w_F}))\nabla_{w_F} x_{w_F} (\nabla_{x_{w_F}}f_{\theta}^i(x_{w_F}) - \nabla_{x_{w_F}}f_{\theta}^y(x_{w_F})) + 0}{\sum_{i=1}^{K}\exp(f_{\theta}^i(x_{w_F}))} \\
					=&\frac{\sum_{i=1}^{K}\exp(f_{\theta}^i(x_{w_F}))\nabla_{w_F} x_{w_F} (\nabla_{x_{w_F}}f_{\theta}^i(x_{w_F}) - \nabla_{x_{w_F}}f_{\theta}^y(x_{w_F}))}{\sum_{i=1}^{K}\exp(f_{\theta}^i(x_{w_F}))}\\
					=&\nabla_{w_F} \mathcal{L}^{\text{CE}}_{\theta}(x_{w_F}, y)
			\end{aligned} \end{equation}
			Finally, we denote $\frac{\sum_{i \neq y}^{K}\exp(f_{\theta}^i(x_{w_F}))}{\sum_{i=1}^{K}\exp(f_{\theta}^i(x_{w_F}))}$ as $r(x_{w_F},y)$.
		\end{proof}

		%

		\section{Proof of Proposition~\ref{prop:affine}}~\label{appendix:max}
		
		
		\begin{proof}
			Let f as the minimizer, e.g., the neural networks after the optimization:
			\begin{equation}
				\begin{aligned}
					f^{\star} & \in \min _{f} \max_i \mathcal{R}(f,S_i) \\
					M^{\star} &=\max_i \mathcal{R}(f,S_i)
				\end{aligned}
			\end{equation}
			Then the optimization can be equivalent to a constrained version:
			\begin{equation}
				\begin{aligned}
					\min_{f, M} & \ \ M \\
					\text { s.t. } & \mathcal{R}(f,S_i) \leq M \text { for all } S_{i} \in \mathcal{S}
				\end{aligned}
			\end{equation}
			with Lagrangian $L(f, M, \lambda)=M+\sum_{S_{i} \in \mathcal{S}} \lambda_{i}\left(\mathcal{R}(f,S_i)-M\right)$. If this optimization problem satisfies KKT condition, then $\exists \lambda_i \ge 0$ with $\nabla_{f} L\left(f^{\star}, M^{\star}, \lambda\right)=0$ such that 
			$$\left.\nabla_{f}\right|_{f=f^{\star}} \sum_{S_{i} \in \mathcal{S}} \lambda_i \mathcal{R}(f,S_i)=0$$
		\end{proof}
		
		\paragraph{Remark} We point out that our conclusion is made under the assumption that the KKT condition holds and the stationary point of $f$ regarding the Lagrangian function can be attained, which normally requires the convexity condition. However, under these assumptions, we can still establish the close correlation between Max AT and Ave AT, indicating they are likely to perform similarly in many cases.
		
		\section{Optimization analysis on the Pareto Adversarial Training and Algorithm}~\label{appendix:pareto_optimization}
		
		We provide the proof of $P$ in the following:
		
		\begin{proof}
			\begin{equation}
				\begin{aligned}
					\sum_{i=0}^{3}\sum_{j=0}^{3}\mathbb{E}(\alpha_{i}\mathcal{L}_i - \alpha_{j} \mathcal{L}_j)^2 =& \sum_{i=0}^{3}\sum_{j=0}^{3}\mathbb{E}((\alpha_{i}\mathcal{L}_i - \mathbb{E}(\alpha_{i}\mathcal{L}_i)) - (\alpha_{j} \mathcal{L}_j - \mathbb{E}(\alpha_{j}\mathcal{L}_j)) + (\mathbb{E}(\alpha_{i}\mathcal{L}_i)- \mathbb{E}(\alpha_{j}\mathcal{L}_j)))^2\\
					=& \sum_{i=0}^{3}\sum_{j=0}^{3}\mathbb{E}((\alpha_{i}\mathcal{L}_i - \mathbb{E}(\alpha_{i}\mathcal{L}_i) - (\alpha_{j} \mathcal{L}_j - \mathbb{E}(\alpha_{j}\mathcal{L}_j))^2 +  (\mathbb{E}(\alpha_{i}\mathcal{L}_i)- \mathbb{E}(\alpha_{j}\mathcal{L}_j))^2 + 0\\
					=&\sum_{i=0}^{3}\sum_{j=0}^{3} (\alpha_{i}^2 \sigma_{ii}+\alpha_{j}^2 \sigma_{jj} - 2\alpha_{i}\alpha_{j}\sigma_{ij}) + (\alpha_{i}^2 \mu_{i}^2+\alpha_{j}^2 \mu_{j}^2 - 2\alpha_{i}\alpha_{j}\mu_{i}\mu_{j})\\
					=& \ 8\alpha^T \text{diag}(\Sigma) \alpha - 2\alpha^T \Sigma \alpha + 8 \alpha^T \text{diag}(\mu \mu^T)  \alpha -2 \alpha^T (\mu \mu^T) \alpha \\
					=& \ \alpha^T (8(\text{diag}(\Sigma)+\text{diag}(\mu \mu^T))-2(\Sigma+\mu \mu^T)) \alpha\\
				\end{aligned}
			\end{equation}
			
		\end{proof}

		\section{Implementation}~\label{appendix:implementation}
		
		\noindent \textbf{Implementation Details.} For MNIST comparison, we train the Simple CNN in \cite{zhang2019theoretically} on MNIST for $100$ epochs. As for the CIFAR-$10$ dataset, we choose the widely used Pre-Act ResNet-$18$ with grouped normalization and trained the network for $76$ epochs. The other details of our implementation on MNIST and CIFAR-$10$ are based on \cite{zhang2019theoretically}, while the implementation on Caltech-$256$ has to refer to \cite{zhang2019interpreting} with $10$ epochs to finetune a pre-trained ResNet-$18$.
		
		\begin{itemize}
			\item \noindent \textbf{PGD Attack.} We apply the widely accepted setting on these three datasets. We set step size as $0.01$, $\epsilon$ as $0.3$ on MNIST while the step size is $0.007$ and $\epsilon$ is $0.031$ on both CIFAR-$10$ and Caltech-$256$ datasets. To evaluate the different levels of robustness, we evaluate PGD attack under $10,20,30,40$ iterations on MNIST and $5,10,15,20$ iterations on CIFAR-$10$ and Caltech-$256$ datasets.
			
			\item \noindent \textbf{Flow-based and RT Attacks.} On MNIST, we set step size $\alpha_{F}$ and $\alpha_{RT}$ as $0.01$ and $0.1$, and choose $\epsilon_{F},\epsilon_{RT}$ as $0.3$. We select $5,10,15,20$ as the attack iterations for the evaluation of both two attacks. On CIFAR-$10$, we set step size $\alpha_{F}$ as $1e-3$ and $\alpha_{RT}$ as $0.05$, and choose $\epsilon_{F},\epsilon_{RT}$ as $0.3,1.0$. We select $3,5,10,15$ as the attack iterations for the evaluation of both two attacks. On Caltech-$256$, we set step size $\alpha_{F}$ as $1e-5$ and $\alpha_{RT}$ as $0.1$, and choose $\epsilon_{F},\epsilon_{RT}$ as $0.3$ and $1.0$ for the two attacks, respectively. We select $3,5,10,15$ as the attack iterations for the evaluation of both two attacks.
			
			\item \noindent \textbf{PGD Adversarial Training.} We choose PGD iterations as 30, 3, and 5 in the PGD adversarial training on MNIST, CIFAR-10, and Caltech-256, respectively. The adversarial attack strength is the same as PGD attacks for each dataset, respectively.
			
			\item \noindent \textbf{Spatial Adversarial Training.} Our integrated spatial adversarial training is based on our proposed integrated spatial attacks that unify both Flow-based and RT-based attacks. We set the iterations as $20$, $5$, $10$ and  on MNIST, CIFAR-$10$ and Caltech-$256$, respectively. Other hyper-parameters are the same as those in their corresponding attacks.
			
			\item \noindent \textbf{Pareto Adversarial Training.} The parameter $r$ is the measure of comprehensive adversarial robustness. We select a sequence of $r$ to train multiple Pareto Adversarial training models. Particularly, on MNIST, we choose $r$ in $[0.2,0.5,0.8,1.0,1.2,1.5,1.8,2.0,2.2]$, and $r$ in $[0.5,1.0,1.25,1.5,2.25,3.0,3.5,4.0]$ on CIFAR-$10$ and Caltech-$256$. Other parameters follow the corresponding methods above, respectively.
		\end{itemize}
		
	\end{appendix}
	
\end{document}